\documentclass[]{article}
\usepackage[letterpaper]{geometry}
\usepackage{mtsummit2021}
\usepackage{times}
\usepackage{url}
\usepackage{latexsym}
\usepackage{natbib}
\usepackage{layout}
\usepackage{graphicx}

\usepackage{color}

\newcommand{\COMMENT}[1]{}
\usepackage{acronym}
\newcommand{\SmarTerp}{SmarTerp}  
 
\begin{document}

\title{\bf Seed Words Based Data Selection for Language Model Adaptation}

\author{\center{\name{\bf Roberto Gretter}, 
        \name{\bf Marco Matassoni}, 
       \name{\bf Daniele Falavigna}\\
        \name{Fondazione Bruno Kessler, Trento, Italy}\\
        \name{(gretter,matasso,falavi)@fbk.eu}\\
        }       
}
\COMMENT{ \author{ 
\center{\name{\bf XXX}, \name{\bf YYY} \name{\bf ZZZ}\\
        \name{aaa, bbb, ccc}\\
        \name{(xxx,yyy,zzz)@aaa.ccc}\\
        }
}
}
\maketitle
\pagestyle{empty}

\begin{abstract}
We address the problem of language model customization in applications where the ASR component needs to manage domain-specific terminology; although current state-of-the-art speech recognition technology provides excellent results for generic domains, the adaptation to specialized dictionaries or glossaries is still an open issue. In this work we present an approach for automatically selecting sentences, from a text corpus, that match, both  semantically and morphologically, a glossary of terms (words or composite words) furnished by the user. 
The final goal is to rapidly adapt the language model of an hybrid ASR system with a limited amount of in-domain text data in order to successfully cope with the linguistic domain at hand; the vocabulary of the baseline model is expanded and tailored, reducing the resulting OOV rate. Data selection strategies based on shallow morphological seeds and semantic similarity via word2vec are introduced and discussed; the experimental setting consists in a simultaneous interpreting scenario, where ASRs in three languages are designed to recognize the domain-specific terms (i.e.\ dentistry). Results using different metrics (OOV rate, WER, precision and recall) show the effectiveness of the proposed techniques.  
\end{abstract}

\section{Introduction}

In this paper we describe an approach to adapt the Language Models (LMs)  used in a system designed to  give help to simultaneous interpreters.  Simultaneous interpreting is a very difficult task that requires a high cognitive effort   especially to correctly translate parts of the source language that convey important pieces of information for the final users. These are: numerals, named entities and technical terms specific of each interpretation session. As an example, a study reported in \cite{fantinuoli2018}
claims that the error rate made by professional interpreters  on the translation of numbers is, on average, equal to 40\%.

This demands for a technology, based on automatic speech recognition (ASR), capable of  detecting, in real time and  with high accuracy, the important information (words or composite terms) of a speech to interpret  and to provide it to a professional interpreter by means of a suitable interface. Therefore, our goal is not to minimise the word error rate (WER) of an audio recording, as usual in ASR applications, instead we aim to maximise the performance of the developed system, in terms of precision, recall and F-measure, over  a set of ``important" terms to recognise, as will be explained in section~\ref{sec:benchmark}. To do this we experimented on a set of data properly labelled by human experts. 

It is worth to point out that this task is different from the usually known ``keywords spotting" task,  since we cannot assume to know in advance the terms to spot inside the audio stream but we can only start from some ``seed" terms belonging to a glossary which is part of the experience of each human interpreter. This demands for further processing modules that: {\em a)} extend, in some way, the given glossary including also  "semantically" similar terms, as will be explained in section~\ref{sec:w2v}, in order to adapt both the dictionary and the language model (LM) employed in the ASR system, and/or  {\em b)} detect along an automatically generated transcription the pieces of information (i.e.\ numerals, named entities, etc) useful to the interpreter.
Actually, the ASR system described below is part of a bigger system that integrates natural language processing (NLP) modules, dedicated to both named entity and numeral extraction, and a user interface specifically designed according to the requirements of professional interpreters. This system, named  
\SmarTerp\footnote{The \SmarTerp\ Project is funded by 
EIT DIGITAL 
under contract n. 
21184}, 
aims to  support the simultaneous interpreters in  various phases of their activities: the preparation of glossaries, automatic extraction and display of the ``important" terms of an interpreting session, post-validation of new entries~\citep{rodriguez2021}.

{\bf Related works. } As previously mentioned spotting known words from audio recordings is a largely investigated task since the beginning of speech recognition technology 
(e.g.\ see works reported 
in ~\cite{bridle73,Rose1990,Weintraub95}. 
Basically all these  approaches used scores derived from acoustic log-likelihoods of recognised words to take a decision of keyword acceptance or rejection. 

More recently, with incoming of neural networks,  technologies have begun to take hold based on deep neural networks \citep{Chen2014}, convolutional neural networks \citep{Sainath2015} and recurrent neural networks \citep{Fernandez2007} to approach keyword spotting tasks. The last frontier is the usage of end-to-end neural architectures  capable of modelling sequences of acoustic observations, such as the one described in~\cite{Yan2020} or the sequence transformer network described in~\cite{berg2021keyword}. 
\COMMENT{
However, as seen above the particular domain application of this work doesn't allow to have a prior knowledge of all of the important terms to detect and, in addition, the NLP modules specialised to text processing need the whole automatic transcription generated by the ASR system to perform both numerals and named entities recognition. To cope with these requirements we decided to include in the ASR language model as much domain information as possible by extracting it from some, possible large, general text corpora.  }

The approach we use for enlarging the dictionary of the ASR system and to adapt the corresponding language model to the application domain is to select and use from  a given, possibly very large and general text corpus, the sentences that exhibit a certain ``similarity" with the terms included in the glossaries furnished by the interpreters. Similarly to the keyword spotting task,  ``term based similarity" represents   a well investigated topic in the scientific community since many years. A survey of approaches can be found in the work reported in~\cite{Vijaymeena2016}. Also for this task the advent of neural network based models has allowed significant improvements both in the word representation, e.g.\ with the approaches described in~\cite{mikolov2013}, and in text similarity measures, e.g.\ as reported in ~\cite{mikolov2014,kareem2019}.

Worth to notice is that in the ASR system used for this  work we do not search for new texts to adapt the LM, instead, as explained in section~\ref{sec:selection}, we select the adaptation texts from the same corpus used to train  the baseline LM. 
Note also that our final goal  is not that to extract  the named entities from the ASR transcripts - this task is accomplished by the NLP modules mentioned above -  instead it consists in providing to the ASR system a LM  more suitable to help the human interpreter of a given event.
Also for ASR system adaptation there is an enormous scientific literature, both related to language models and to acoustic models adaptation;  here we only refer some recent papers:~\cite{song-etal-2019-chameleon} for LM adaptation and ~\cite{bell2021} for a review of acoustic model adaptation approaches, especially related to neural models.

\section{Automatic selection of texts}
\label{sec:selection}

Usually a Language Model (LM) is trained over huge amounts of text data in a given language, e.g.\ Italian. During the training phase, a fixed lexicon is selected - typically the N most frequent words in the text - and millions or billions of n-grams are stored to give some probability to any possible word sequence. This process allows to build a somehow generic LM, capable to represent the language observed in the text.

However, interpreters often need to specialise their knowledge on a very specific topic, e.g.\  dentistry. In this case, they also have to quickly become experts in that particular field. We could say that they need to adapt their general knowledge to that field: this means that, before the event, they have to collect material about that topic, study it, prepare and memorise a glossary of very specific technical terms together with their translations.

The same process holds for an ASR system: it can perform in a satisfactory way in a general situation, but it may fail when encountering technical terms in a specific field. So, it has to be adapted, both in terms of lexicon (it may be necessary to add new terms to the known lexicon) and in terms of word statistics for the new terms.

In the \SmarTerp\ project we are going to explore different adaptation procedures and describe in this paper  our preliminary work in this direction. At present, we hypothesise that an interpreter could provide some text and the ASR system will be able to adapt to the corresponding topic in a short time (some hours on a medium computer). 
This short text could range from a few words to a quite large set of documents that identify that particular topic, depending on the expertise and the attitude of the interpreter. Here are some possibilities: 
\begin{itemize}
\itemsep-0.3em
\item just a few technical words;
\item a glossary of terms, maybe found with a quick internet search;
\item a glossary of technical terms with translations, maybe built over the years by an expert interpreter;
\item a set of technical documents, in the desired language.
\end{itemize}
In a very near future, in \SmarTerp\ a pool of interpreters will be engaged in simulations where they have to provide data that, in a complete automatic way (i.e.\ without the intervention of some language engineer), will adapt the ASR system for a particular topic.
In this work we are testing some tools and procedures in order to provide them some possible solutions, assuming that at least some small text (i.e.\ a glossary, or even a few words) will be available. From this small text we will derive some {\em seed words} that will be used, in turn, both to update the dictionary of the ASR system and to select LM adaptation  texts from the available training corpora (see Table~\ref{tab:LMdata}). 
In detail, we implemented the following procedures (although some of them were not used in the experiments described in this paper):
\begin{itemize}
\itemsep-0.3em
\item selection of {\bf seed words}, i.e.\ technical words that characterise the topic to be addressed; they are simply the words, in the short text provided by the interpreter, that are not in the initial lexicon, composed of the most frequent N words of that language (128 Kwords, in this paper).
\item optional enlargement of the set of {\bf seed words}, either by exploiting shallow morphological information or using neural network approaches like word2vec \citep{mikolov2013}.
\item selection of {\bf adaptation text}, i.e.\ text sentences in the text corpus that contain at least one of the seed words. Note that we hypothesise not to have new texts belonging to the topic to be addressed, that could be directly used for LM adaptation.
\item compilation of an {\bf adapted lexicon} and of an {\bf adapted LM}, obtained exploiting the adaptation text.
\end{itemize}

\subsection{Shallow morphological seed words enlargement}

Each initial seed word is replaced by a regular pattern which removes the ending part, to find similar words in the complete dictionary of the corpus. Possible parameters are: $N_M$, maximum number of similar words retained for each seed; $L_M$, minimal length of a seed pattern to be considered valid (too short patterns are useless or even dangerous).

\subsection{Semantic similarity based approach}
\label{sec:w2v}
Each initial seed word is fed to a pretrained  neural skipgram model (word2vect, see http://vectors.nlpl.eu/repository), which  returns an embedded representation of words. Then, the $N$ more similar words are computed using the cosine distance between couples of words embeddings. The process can be iterated by feeding word2vec with every new similar word obtained. Possible parameters are: $N_W$, number of retained words from each term; $I_W$, number of iterations: typically 1, or 2 in case of a very short list of initial seeds.

\subsection{Selection of adaptation text}

Given a final set of seed words, the huge text corpus is filtered and every document containing at least one seed word, not contained in the (128K) initial lexicon, is retained. One parameter of the filter - not used in this work - is the number of words forming the context around every seed word in a document. This may be useful to avoid to include in the adaptation corpus  useless pieces of texts, due to the fact that every line in the training corpora (newspaper or Wikipedia, title or article) is considered a document, containing  from few words to tens (even hundreds in few cases) of Kwords. Note that the selection of the adaptation text is largely responsible of the lexicon enlargement (up to 250 Kwords, see Table~\ref{tab:results}), since the number of seed words resulted to be, in our preliminary experiments, always below 4 Kwords.

\section{ASR systems}

The ASR system is based on the popular Kaldi toolkit~\citep{kaldi}, that provides optimised modules for hybrid architectures; the modules support arbitrary phonetic-context units, common feature transformation,  Gaussian mixture and neural acoustic models, n-gram language models and on-line decoding. 


\subsection{Acoustic models}
The acoustic models are trained on data coming from CommonVoice~\citep{ardila2020} and Euronews transcriptions~\citep{gretter2014}, using a standard {\em chain} recipe based on lattice-free maximum mutual information (LF-MMI) optimisation criterion ~\citep{povey2016}. In order to be more robust against possible variations in the speaking rate of the speakers, the usual {\em data augmentation} technique for the models has been expanded, generating time-stretched versions of the original training set (with factors $0.8$ and $1.2$, besides the standard factors $0.9$ and $1.1$).

Table~\ref{tab:audio_data} summarises the characteristics of the audio data used for the models in the three working languages considered in the project.

\begin{table}[t]
\begin{center}
\begin{tabular}{|l|c|c|c|c|}
\hline
Language & CV (h:m) & EuroNews (h:m) & Total Speakers   & Running words    \\
\hline 
English  &   781:47  & 68:56 & 35k  & 5,742k \\
Italian  &   148:40  & 74:22 &  9k  & 1,727k \\
Spanish  &   322:00  & 73:40 & 16k  & 2,857k \\
\hline
\end{tabular}
\caption{Audio corpora for AM training}
\label{tab:audio_data}
\end{center}
\end{table}

\begin{table}[tbh]
\begin{center}
\begin{tabular}{|l|c|c|r|r|}
\hline
Language & Lexicon size & Total running words  & Internet News & Wikipedia 2018  \\
\hline 
English  & 9.512.829    & 3790.55 Mw   & 1409.91 Mw       & 2380.64 Mw \\
Italian  & 4.943.488    & 3083.54 Mw   & 2458.08 Mw       &  625.46 Mw \\
Spanish  & 4.182.225    & 2246.07 Mw   & 1544.51 Mw       &  701.56 Mw \\
\hline
\end{tabular}
\caption{Text corpora for training the LMs for ASR in the three \SmarTerp\ languages. Mw means millions of running words.}
\label{tab:LMdata}
\end{center}
\end{table}  

\subsection{Language models and Lexica}

Text corpora that can be used to train LMs for the various languages are described in Table~\ref{tab:LMdata} and derive both from Internet news, collected from about 2000 to 2020, and from a Wikipedia dump; their corresponding total lexica amount to several millions of words (from 4 to 10) for every language. It has to be clarified that, being the original texts definitely not clean, most of the low frequency words are in fact non-words (typos, etc.). For practical reasons, the size of the lexicon used in the ASR usually ranges from 100 to 500 Kwords.

The baseline language models are trained using the huge corpora described in Table~\ref{tab:LMdata}; the adaptation set is selected from the same huge corpora. After the selection stage, the resulting trigrams are computed and a mixed LM is built and then pruned to reach a manageable size. 
The adapted LM probabilities are efficiently derived using the approach described in~\cite{federico2001}   by interpolating the frequencies of trigrams of the  background (i.e.\ non adapted) LM  with the corresponding frequencies  computed on the adaptation text.

The most frequent 128Kwords of the corpus are retained; all the words of the adaptation set are then included in the corresponding lexicon. 

\section{Description of \SmarTerp\ multilingual benchmark}
\label{sec:benchmark}

As mentioned above, in \SmarTerp\ we prepared 
benchmarks for the 3 languages of the project: English, Italian, Spanish. For each language, a number of internet videos having Creative Commons  licence were selected, in order to reach at least 3 hours of material on a particular topic, dentistry. 
Table~\ref{tab:benchmark} reports duration and number of words of the benchmarks. Data were collected, automatically transcribed and manually corrected\footnote{We are really grateful to 
Susana Rodr\'iguez,  
who did the manual check for all the languages.} using Transcriber\footnote{http://trans.sourceforge.net/}, a tool for segmenting, labelling and transcribing speech. 
In addition to time markers and orthographic transcription of the audio data, we decided to label with parenthesis Important Words (IWs), which represent content words that are significant for the selected domain (i.e.\ dentistry) and are a fundamental part of the desired output of the automatic system. As only one annotator labelled IWs, it was not possible to  compute annotators' agreement for this task. We will address this issue in future works.
\begin{table}[bh]
\begin{center}
\begin{tabular}{|l|c|c|c|c|c|}
\hline
language & recordings &  raw  &  transcribed  & running  & running \\
 &  &   duration &   duration &  words &  IWs\\
\hline
English  &  5         & 04:02:34      & 03:03:06     & 28279      & 3343 \\
Italian  & 33         & 05:29:34      & 04:10:31     & 31001      & 4560 \\
Spanish  & 13         & 03:09:53      & 03:01:59     & 25339      & 3351 \\
\hline 
\end{tabular}
\caption{Benchmarks collected and annotated in \SmarTerp.}
\label{tab:benchmark}
\end{center}
\end{table}

\begin{figure}
    \centering
    \includegraphics[scale=0.5]{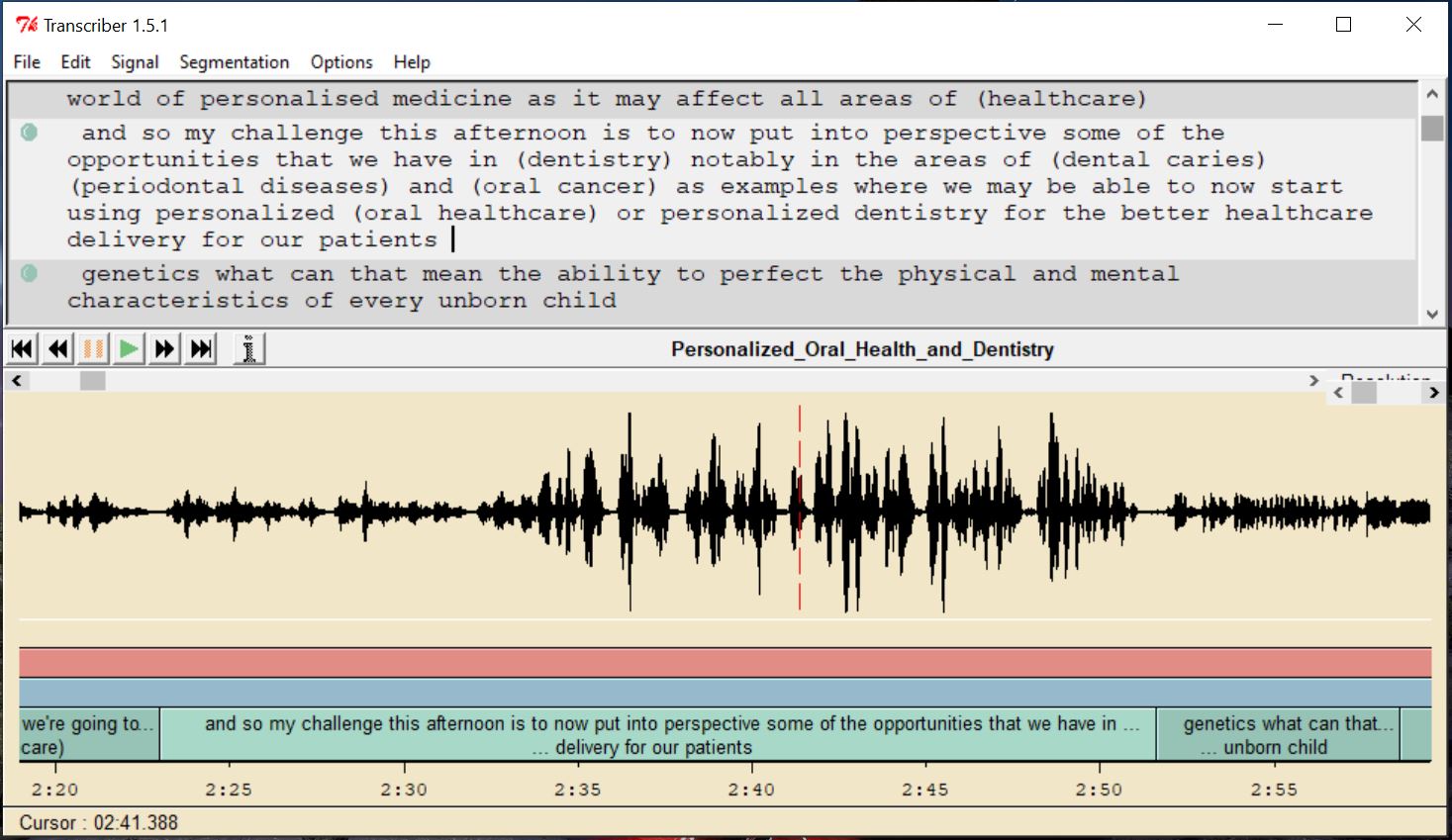}
    \caption{Screenshot of Transcriber, a tool used to manually transcribe the \SmarTerp\ benchmark. In the highlighted segment, IWs are in parentheses.}
    \label{fig:transcriber}
\end{figure}

Figure~\ref{fig:transcriber} shows a screenshot of Transcriber, where some IWs are highlighted: (dentistry), (dental caries), (periodontal diseases), (oral cancer). In the benchmarks, phrases composed up to 6 words were identified as IWs.

\subsection{IW normalization}

In order to be able to consistently evaluate the performance of the system in terms of IWs, and considering that it was impossible to pre-define a fixed set of IW patterns, we decided to implement a procedure that automatically processed the whole benchmark. It consisted of the following basic steps, applied independently for every language:
\begin{enumerate}
\itemsep-0.3em
    \item identification of all manually defined IWs in the benchmark;
    \item reduction to a minimum set of IWs, by removing ambiguities. Given that A, B, C, etc. are single words, some cases are:
    \begin{itemize}
    \itemsep-0.3em
    \item if exist (A), (B) and (A B), then the IW (A B) is removed - will be replaced by (A) (B);
    \item if exist (C), (D E) and (C D E), then the IW (C D E) is removed;
    \item note however that if exist (C), (D E) and (D C E), nothing can be removed.
    \end{itemize}
\item regeneration of the benchmark, by applying the following steps:
    \begin{enumerate}
    \itemsep-0.3em
    \item remove all round brackets;
    \item considering the minimum set of IWs, apply new brackets at every IW occurrence, starting from the longest IWs and ending with the one-word IWs;
    \item in order to evaluate Precision, Recall and F-measure of IWs, remove all words not inside brackets.
    \end{enumerate}
    
\end{enumerate}
Note that some IWs originally present in the benchmark, although legitimate, could not appear in the final version of the benchmark: suppose that the only occurrence of (B) alone is in the context A (B) and also the IW (A B) exist: after the regeneration of the benchmark, both cases will result (A B).

\begin{table}
\footnotesize{
\begin{center}
\begin{tabular}{|l|p{10cm}|}
\hline
REF &          the most of {\bf them} referred from (pulmonary specialist) {\bf (ENTs)} (paediatricians) {\bf let's let Boyd try} nothing {\bf else} \\
ASR & {\bf in} the most of {\bf my}   referred from (pulmonary specialist) {\bf ian}    (paediatricians) {\bf was  led  by  tried} nothing \\
ALIGNMENT & I\_in S\_them\_my S\_ENTs\_ian S\_let's\_was S\_let\_led S\_Boyd\_by S\_try\_tried D\_else (Sub= 6 Ins= 1 Del= 1 REF=16) \\
WER & 50.00\% [ 100 * (6 +1 +1) / 16 ] \\ \hline
IW-REF & (pulmonary\_specialist) {\bf (ENTs)} (paediatricians) \\
IW-ASR & (pulmonary\_specialist)        (paediatricians) \\
P / R / F &  Precision 1.00 [ 2 / 2 ] / Recall 0.67 [ 2 / 3 ] / F-Measure  0.80  \\ \hline
Isol-IW-REF & (pulmonary) (specialist) {\bf (ENTs)} (paediatricians) \\
Isol-IW-ASR & (pulmonary) (specialist)        (paediatricians) \\
P / R / F &  Precision 1.00 [ 3 / 3 ] / Recall 0.75 [ 3 / 4 ] / F-Measure  0.86  \\ \hline
\end{tabular}
\caption{Evaluation metrics on a sample of the English benchmark: WER over the whole text; Precision, Recall, F-measure over both the IWs and the Isolated-IWs. ASR errors are highlighted in bold. IWs are those in parentheses.}
\label{tab:metrics}
\end{center}
}
\end{table}

After the application of this algorithm, a consistent version of the benchmark was obtained. By applying the same regeneration steps to the ASR output, a fair comparison was possible, considering only the IWs. We could also consider different metrics, either by considering each IW as a single item (despite the number of words that compose it) or by considering separately each word that compose the IWs (henceforth Isol-IW).
Standard evaluation of ASR output is Word Error Rate (WER), resulting from a word-by-word alignment between reference text (REF) and ASR output (TEST). In detail, WER is the percentage of substitution, insertions and deletions over the number of REF words. In \SmarTerp, however, it could be more useful to concentrate on the IWs only, and to consider Precision, Recall and F-Measure as primary metric.
The example in Table~\ref{tab:metrics} shows the different metrics used in this work.

\subsection{Preliminary analysis}

\begin{figure}[bh]
    \centering
    \includegraphics[scale=0.5]{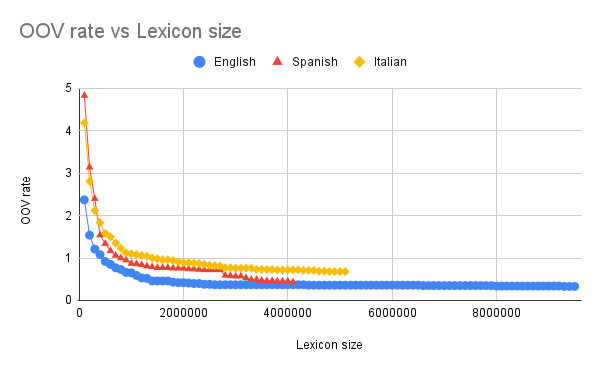}
    \caption{OOV rate of the \SmarTerp\ benchmarks against lexicon size for the 3 languages.}
    \label{fig:OOVLex}
\end{figure}

Figure~\ref{fig:OOVLex}  reports OOV rate of the \SmarTerp\ Benchmark for different values of the lexicon size, computed on all the available text data described in  Table~\ref{tab:LMdata}. 
An inspection of OOV words was done for the Italian language, in order to better understand how the OOV words are distributed among different classes. 
With respect to the 128 Kwords lexicon, we had that the Italian benchmark 
is composed of $31001$ running words, of which $1089$ are OOV (corresponding to 3.51\% OOV rate). The number of different OOV words was 474, manually classified as follows:
\begin{itemize}
\itemsep-0.3em
\item {\bf 190 Morpho}: morphological variations of common words (e.g.\ allunghiamo, distinguerle, divideremo - {\it we lengthen, distinguish them, we will divide});
\item {\bf 181 Tech}: technical terms, that will be part of IWs so it is extremely important to keep their number as low as possible
(e.g.\ bruxismo, implantologia, parodontopatici  - {\it bruxism, implantology, periodontal disease});
\item {\bf 34 Errors}: words that should not be here and will be fixed soon: numbers in letters, wrong tokenization (e.g.\ cinque, computer-assistita, impianto-protesica, l'igiene  - {\it five, computer-assisted, implant-prosthetic, the hygiene});
\item  {\bf 28 English}: terms in English, often they are technical terms and should be recognized (e.g.\  osteotomy, picking, restaurative, tracing);
\item {\bf 20 Names}: proper names of people, firms or products (e.g.\ claronav, davinci, hounsfield, navident);
\item {\bf 10 Latin}: latin words (e.g.\ dolor, restitutio, tumor - {\it pain, restoration, swelling});
\item {\bf 8 Acronyms}: (e.g.\ t-test, mua, d3, d4);
\item {\bf 3 Foreign}: pseudo-foreign words that need particular care for pronunciation (e.g.\ customizzata, customizzati, matchare - {\it Italian neologisms from English custom, match}).
\end{itemize}
\begin{table}[bt]
\footnotesize{
\begin{center}
\begin{tabular}{|rl|rl|rl|}
\hline
\multicolumn{2}{|c|}{allunghiamo} &
\multicolumn{2}{ c }{distinguerle} &
\multicolumn{2}{|c|}{divideremo} \\
\multicolumn{2}{|c|}{{\it we lengthen}} &
\multicolumn{2}{ c }{{\it distinguish them}} &
\multicolumn{2}{|c|}{{\it we will divide}} \\
\hline 
 10355 &allunga         & 12118 &distingue       &  7273 &divide          \\
 12657 &allungare       & 12493 &distinguere     &  7931 &dividendo      \\
 17187 &allungato       & 20484 &distinguono     & 12286 &dividere        \\
 18040 &allungo         & 26323 &distinguo       & 14127 &dividendi      \\
 20126 &allungamento    & 34366 &distinguersi    & 15601 &dividono        \\
 23870 &allungano       & 52496 &distinguendosi  & 27370 &dividersi      \\
 25749 &allungata       & 56673 &distingueva     & 43165 &divideva        \\
 35514 &allungando      & 60858 &distinguerlo    & 59956 &dividerà      \\
 40996 &allungate       & 61213 &distinguendo    & 61370 &dividerci      \\
 42540 &allungati       & 67741 &distinguibili   & 62319 &divideranno    \\
 43104 &allungarsi      & 75608 &distinguerla    & 63369 &dividendosi    \\
 60394 &allunghi        & 77105 &distinguibile   & 68113 &dividevano      \\
 98044 &allungherà      & 79891 &distinguevano   & 80977 &dividerli      \\
106019 &allungava       & 91152 &distinguerli    & 84294 &dividend        \\
120007 &allungandosi    &115236 &distinguiamo    & 91609 &divida          \\
126079 &allungherebbe   &116550 &distingua       & 97706 &dividiamo      \\
      &                 &119097 &distinguerà     &121708 &dividerlo      \\
\hline
\end{tabular}
\caption{Morphological variations of OOV words, known in the 128 Kwords lexicon, along with their position in the lexicon.}
\label{tab:morpho}
\end{center}
}
\end{table}  
Tech, English, Names, Latin and Foreign will deserve a particular attention in future studies, because they are important for the domain. 
Errors will be fixed and should disappear; 
Acronyms should be recognized as subwords (e.g., d3 as d 3).
Morpho will probably be misrecognized as another morphological variation of the same stem, present in the active dictionary, which in this domain is not considered a critical error. Note that a single verbal stem in Italian can generate up to 300 different words in Italian, including clitics. In Table~\ref{tab:morpho} you can see the morphological variations of the 3 terms of the class Morpho reported above which are present in the 128 Kwords lexicon.
\begin{figure}[th]
    \centering
    \includegraphics[scale=0.45]{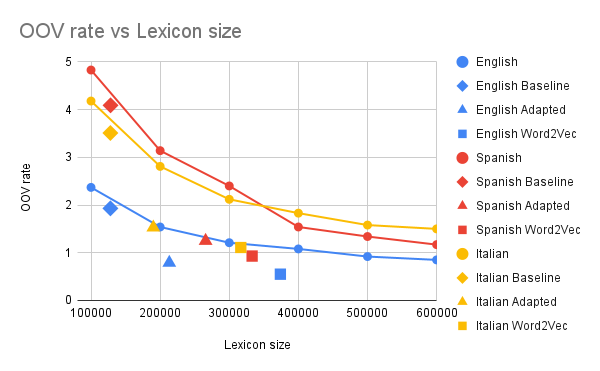}
    \caption{OOV rate of the \SmarTerp\ benchmarks against lexicon size for the 3 languages, for all the experiments and languages.}
    \label{fig:OOVLexExpe}
\end{figure}
\section{Experiments and results}

Since several approaches can be employed to obtain, enlarge and use the seed words (e.g.\  based on texts distance, texts semantic similarity, etc) we consider the following indicators that allow to measure their effectiveness on the benchmarks collected and manually transcribed within the \SmarTerp\ project. 
\begin{itemize}
\itemsep-0.3em
\item Seeds: number of seed words, used to extract the adaptation text:
\item Out Of Vocabulary rate (OOV rate): it is the percentage of unknown words in the benchmark, with respect to the lexicon. OOV words cannot be part of the output of the ASR, hence they will be certainly errors. We should try to get a low OOV rate without the lexicon size growing too much;
\item Lexicon size: total number of active words in the adapted LM;
\item  Word Error Rate (WER): it measures the percentage of errors made by the ASR;
\item  Precision, Recall, F-Measure over the set of Important Words (IWs) that were defined. 
\end{itemize}

The following experiments were carried out for each of the three languages:
\begin{itemize}
\itemsep-0.3em
\item 
 {\bf Baseline}: the initial 128Kwords lexicon and the LM trained on the whole corpus, without any adaptation; 
\item {\bf Adapted}: LM adapted starting from seed words coming from a dental glossary (normally 2-3 pages of text, resulting into some hundreds of seeds), found with a quick search in internet for terms like ``dental glossary'' (e.g.\ https://bnblab.com/intro/terminology).
\item  {\bf Word2Vec}: LM adapted using seed words obtained from 5 initial seed words, applying two iterations ($I_w=2$) of the procedure based on semantic similarity and retaining, for each term,  $N_w=40$ words, obtaining $\sim 3000$ seed words. The 5 magic words\footnote{Many thanks to Susana Rodr\'iguez  
 for the translations of the magic words from Italian} were:
\begin{itemize}
\itemsep-0.3em
\item  {\bf English}: tartar, filling, caries, tooth, dentist
\item  {\bf Italian}: tartaro, otturazione, carie, dente, dentista
\item  {\bf Spanish}: sarro, relleno, caries, diente, dentista
 \end{itemize}
\end{itemize}
Figure~\ref{fig:OOVLexExpe}  reports OOV rate of the \SmarTerp\ benchmark for different values of the lexicon size for each experiment, along with the initial part of the curve of Figure~\ref{fig:OOVLex}. It should be noted that, for every language,  Baseline is along the initial curve, while both Adapted and Word2Vec are well below it. 
For all languages, Adapted has a Lexicon size which is in between Baseline and Word2Vec. This is due to an initial choice of the parameters described in Section~\ref{sec:selection}: by changing the parameters, a cloud of values could be generated instead of a single point. In fact, in this work we report only initial experiments and future efforts will be devoted to a parameter optimization. In any case, the Lexicon size is directly related to the number of seeds and on the size of the adaptation text, which plays a very important role in the adaptation stage.

Table~\ref{tab:results} reports preliminary results 
on the three benchmarks, for all the experiments. Together with the number of obtained seed words, OOV rate and Lexicon size, we report WER computed on all the uttered words (including functional words, which are useless for this task), and Precision/Recall/F-measure computed both on IWs and Isol-IWs: since they represent the most technically significant words in the domain, they are more related to the output desired by interpreters. 
It is worth noting that, with respect to Baseline, both the Adapted and Word2Vec  systems are effective for all of the three languages and for all the considered metrics. Word2Vec performs slightly better than Adapted, but this can be due to the initial value of the parameters that bring to more seeds and to a bigger Lexicon size. 
Low WER for English is partly due to a scarce audio quality in the recordings, that mainly affects functional words: this explains the English high precision, which is computed on IWs only.

\begin{table}[thb]
\begin{center}
\begin{tabular}{|l|r|c|c|c|c|c|}
\hline
  & Seeds & Lex size& OOVrate& WER     & IW P / R / F      & Isol-IW P / R / F   \\ \hline 
 Eng BL   &    0 & 128041 &    1.93\%    & 26.39\% & 0.90 / 0.61 / 0.73 & 0.96 / 0.59 / 0.73 \\
 Eng ada  &  257 & 213237 &    0.79\%    & 23.34\% & 0.92 / 0.73 / 0.81 & 0.97 / 0.71 / 0.82 \\
 Eng w2v  & 2999 & 373956 &    0.55\%    & 23.86\% & 0.93 / 0.72 / 0.81 & 0.97 / 0.70 / 0.81 \\ \hline
 Ita BL   &    0 & 128009 &    3.51\%    & 15.14\% & 0.88 / 0.67 / 0.76 & 0.95 / 0.67 / 0.79 \\
 Ita ada  &  213 & 190126 &    1.53\%    & 11.73\% & 0.96 / 0.84 / 0.89 & 0.98 / 0.82 / 0.90 \\
 Ita w2v  & 3527 & 316679 &    1.11\%    & 11.28\% & 0.96 / 0.85 / 0.90 & 0.99 / 0.84 / 0.91 \\ \hline
 Spa BL   &    0 & 128229 &    4.09\%    & 22.60\% & 0.86 / 0.56 / 0.68 & 0.93 / 0.56 / 0.69 \\
 Spa ada  &  673 & 265764 &    1.25\%    & 17.74\% & 0.95 / 0.76 / 0.85 & 0.98 / 0.75 / 0.85 \\
 Spa w2v  & 3207 & 333072 &    0.93\%    & 17.31\% & 0.95 / 0.79 / 0.86 & 0.98 / 0.78 / 0.87 \\ \hline
\end{tabular}
\caption{Preliminary results for Baseline (BL), Adapted (ada) and Word2Vec (w2v) systems. Both WER on all words and Precision/Recall/F-measure on composite and isolated IWs are reported.}
\label{tab:results}
\end{center}
\end{table}  

\section{Conclusions}

We described two different approaches for extending the dictionary of an ASR system in order to detect important terms from technical speeches, namely dental reports, to be translated by simultaneous professional interpreters. The two approaches consist in extracting adaptation text from a huge set of text data, starting from some seed words. In the first one, seed words come from a given glossary.
The second one is based on the application of a text similarity measure to an initial (very small) set of $5$ seed words. After the application of the selection procedures we adapted the language models used in the ASR system employed in a computer assisted interpretation  (CAI) system under development and we proved the effectiveness on the approaches in terms of different evaluation metrics.

\small

\bibliographystyle{apalike}
\bibliography{mtsummit2021,mybib}

\end{document}